\documentclass[sigconf]{acmart}

\renewcommand\footnotetextcopyrightpermission[1]{} 
\usepackage[utf8]{inputenc}
\usepackage{xcolor}
\usepackage{xspace}
\usepackage{multirow}
\usepackage{dblfloatfix}
\usepackage{subcaption}

\usepackage[sort&compress,capitalize,noabbrev,nameinlink]{cleveref}
\creflabelformat{equation}{#2#1#3}

\newcommand\Tstrut{\rule{0pt}{2.6ex}}         
\newcommand\Bstrut{\rule[-0.8ex]{0pt}{0pt}}   

\newcommand*{\eg}{e.g.\@\xspace}
\newcommand*{\ie}{i.e.\@\xspace}

\AtBeginDocument{%
  \providecommand\BibTeX{{%
    \normalfont B\kern-0.5em{\scshape i\kern-0.25em b}\kern-0.8em\TeX}}}

\setcopyright{none}

\acmSubmissionID{2}

\begin{document}

\title{MonoComb: A Sparse-to-Dense Combination Approach for Monocular Scene Flow}

\author{René Schuster}
\email{rene.schuster@dfki.de}
\author{Didier Stricker}
\email{didier.stricker@dfki.de}
\affiliation{%
  \institution{German Research Center for Artificial Intelligence -- DFKI}
}

\author{Christian Unger}
\email{christian.unger@bmw.de}
\affiliation{\institution{BMW Group}}

\renewcommand{\shortauthors}{Schuster et al.}

\begin{abstract}
	Contrary to the ongoing trend in automotive applications towards usage of more diverse  and more sensors, this work tries to solve the complex scene flow problem under a monocular camera setup, i.e. using a single sensor. Towards this end, we exploit the latest achievements in single image depth estimation, optical flow, and sparse-to-dense interpolation and propose a monocular combination approach (MonoComb) to compute dense scene flow. MonoComb uses optical flow to relate reconstructed 3D positions over time and interpolates occluded areas. This way, existing monocular methods are outperformed in dynamic foreground regions which leads to the second best result among the competitors on the challenging KITTI 2015 scene flow benchmark.
\end{abstract}

%

\keywords{Combination, interpolation, monocular, neural networks, scene flow}

\begin{teaserfigure}
	\includegraphics[width=0.98\textwidth]{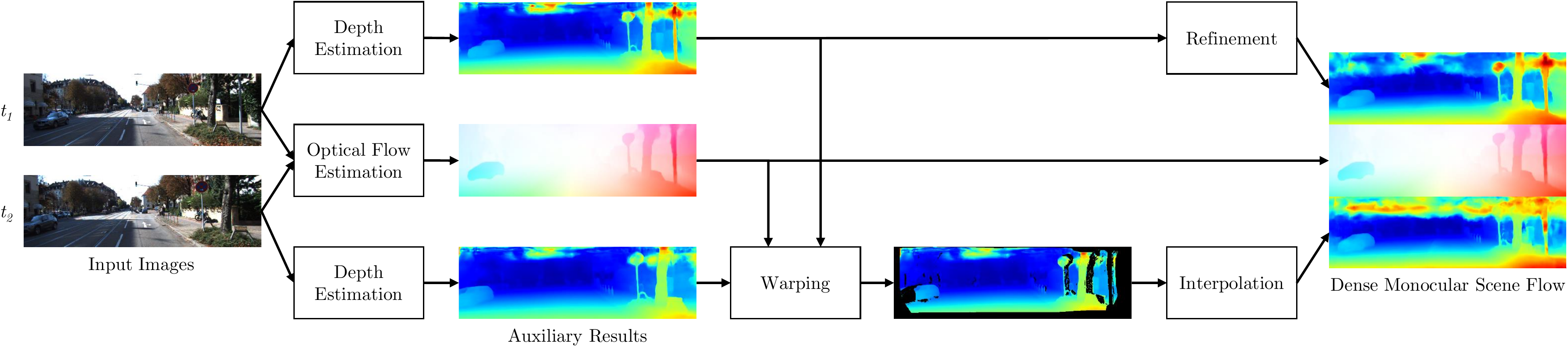}
	\caption{Overview of our pipeline for monocular dense scene flow estimation.}
	\Description{A block diagram showing the structure and data flow of the proposed approach.}
	\label{fig:overview}
\end{teaserfigure}

\maketitle
\pagestyle{plain}

\section{Introduction} \label{sec:intro}
One current trend for many applications in assisted or autonomous driving is the utilization and fusion of as many sensors as available.
As a result, certain approaches have increasing requirements on the hardware in a product.
A different strategy is to solve the same problems based on input from less sensors, which allows to have the same functionality at lower cost.
This becomes possible by adding constraints or assumptions to the formulation of the problem, by technical progress, or by relying on more or other visual cues.

In this work, the problem at hand is the estimation of dense 3D scene flow.
Scene flow can be considered the 3D equivalent of 2D optical flow, which is achieved by additional 3D reconstruction of the scene.
In most conventional approaches, a pair (or sequence) of stereo images is used as input.
The stereo view allows for the 3D reconstruction, while the images at different points in time provide a motion cue and the temporal information.
Recently, methods that operate solely on point clouds from LiDAR sensors were proposed \cite{liu2019flownet3d,behl2019pointflownet}. These approaches however are less dense compared to the standard resolution of images.
Therefore, the fusion of cameras and LiDAR have been considered, either in the stereo case \cite{battrawy2019lidar} or in the monocular case \cite{rishav2020deeplidarflow}, where the LiDAR measurements replace the stereo dependency for 3D reconstruction.

However, following the strategy to reduce the sensor input, we propose a method to estimate scene flow in the purely monocular case.
While a single RGB camera does not provide a geometric cue as in stereo cameras and is also unable to measure depth directly as a LiDAR or RGB-D camera is, this setup poses scene flow estimation as a much more difficult problem. 
To our rescue, latest developments in single image depth estimation have proven that absolute depth can be reconstructed from a single view point \cite{godard2017unsupervised,fu2018dorn,godard2019digging,ochs2019sdnet,lee2019bts}.
This is possible by relying on depth cues like the monocular motion parallax, defocus blur by the field of depth, linear perspective, texture gradients, or the relative size of known objects.
%
%

We exploit the progress in the field of single image depth estimation and use it within the combination framework for scene flow \cite{schuster2018combining} where geometry and temporal image correspondences are estimated separately and are later combined to obtain full 3D scene flow.
Towards that end, we propose to 1.) estimate depth from two single images, 2.) estimate optical flow between these two images, 3.) correlate corresponding 3D points to estimate 3D motion, and 4.) interpolate gaps due to occlusions to obtain a dense result.
The overview of these steps is given in \cref{fig:overview}.
Since our combination approach operates in a monocular camera setting, we term the proposed method \textit{MonoComb}.

\section{Related Work} \label{sec:related}
In the very beginning of scene flow estimation the problem has already been formulated for a monocular camera \cite{vedula2005three}, however with a strong dependency on multiple views.
Shortly after, stereo cameras were used to provide a better geometric cue for depth estimation \cite{huguet2007variational,wedel2008efficient,vogel2013piecewise}.
A parallel evolution was using RGB-D cameras to measure the geometry of the scene directly \cite{herbst2013rgb,hornacek2014sphereflow}.
As mentioned earlier, the emerging use of laser scanners in vehicles has driven the development of scene flow algorithms in the point cloud domain \cite{liu2019flownet3d,behl2019pointflownet,wang2020flownet3d++}.
And lastly, camera and LiDAR information has been fused to solve the problem of scene flow estimation \cite{battrawy2019lidar,rishav2020deeplidarflow}.

Opposed to all these methods, our work presents a method to estimate scene flow from just two consecutive images from a monocular camera.
There are a few approaches considering the same setup.
In Mono-SF \cite{brickwedde2019mono}, pixel-wise depth distributions are estimated for both images and then used in a probabilistic optimization framework to estimate rigidly moving, planar segments for the scene. Though the superpixel segmentation provides a strong regularization and high accuracy, the assumptions of rigidity and planarity introduce errors depending on the granularity of the segmentation. Also, the optimization is computational heavy.
Self-Mono-SF \cite{hur2020self} uses an adaptation of PWC-Net \cite{sun2018pwc} that is trained in an self-supervised manner to jointly estimate 3D position and 3D flow at a quarter resolution.
This methods is able to achieve competitive results after supervised fine-tuning, but the purely unsupervised version lags behind.
Lastly, OpticalExpansion (OE) \cite{yang2020upgrading} exploits the assumption that most of the change of the projected size of objects is due to the change in distance to the observer.
Therefore the authors argue that the relative motion-in-depth can be estimated directly together with optical flow. In conjunction with an estimate for the depth in one frame, the depth at the second frame can be reconstructed to obtain full scene flow.

Since most of the previous work in monocular scene flow estimation -- as well as our proposed approach -- rely on single image depth estimation, we discuss a part of state-of-the-art in this area as well.
LRC \cite{godard2017unsupervised} is a self-supervised approach that uses an encoder-decoder network architecture and trains with a photometric loss and consistency between left and right view of a stereo camera.
The approach was later refined in MonoDepth2 \cite{godard2019digging}.
DORN \cite{fu2018dorn} proposes to use a ordinal regression loss instead of a regular regression loss to train a network for single image depth estimation. 
This formulation is closer to monocular depth cues, where it is often easier to order the depth of regions instead of regressing the absolute depth of each point.
BTS \cite{lee2019bts} represents state-of-the-art across different data sets. The idea of BTS is to fuse depth estimates from multiple scales at full resolution using local planar guidance for up-sampling.

Finally, very related work is the combination approach for stereo scene flow estimation in \cite{schuster2018combining,schuster2018dense}.
Though the sensor setup is different from ours, the ideas are similar: Separate the complex problem of scene flow estimation into several (solved) sub-problems.
In the work by \cite{schuster2018combining}, this concept yields non-dense scene flow, due to occlusion.
The work of \cite{schuster2018dense} solved this limitation by applying the interpolation method of SceneFlowFields \cite{schuster2018sceneflowfields} to the non dense result.
Since our objective is dense scene flow as well, we also apply an interpolation method to fill in the gaps after warping.
However in contrast to these previous approaches, our initial depth estimates are obtained using a single image from a monocular camera.

\section{Monocular Combination Approach} \label{sec:method}
We propose the following method to obtain dense scene flow from a single monocular image pair $I_1$ and $I_2$ at time steps $t_1$ and $t_2$.
First, we use an off-the-shelf single image depth estimator to predict a dense depth map for each of the images, $D_1$ and $D_2$.
We also predict dense optical flow $\mathbf{u} = (u,v)^T$ from the first to the second image with an auxiliary optical flow estimator.
The optical flow result is directly used within our combined scene flow, and further to estimate the (non-dense) change in depth, by warping $D_2$ towards the reference frame at $t_1$.
At this point, a non-dense scene flow is achieved.
Lastly, we interpolate the gaps which originated during warping to reconstruct the dense scene flow.


\subsection{Auxiliary Depth and Optical Flow Estimation} \label{sec:method:auxiliary}
In principle, any methods for depth and optical flow estimation can be used. The performance of the auxilliary methods directly influence the quality of the final scene flow result. Therefore, we experiment with the state-of-the-art in optical flow VCN \cite{yang2019volumetric} and HD3 \cite{yin2019hierarchical} and BTS \cite{lee2019bts} for single image depth estimation.

While for optical flow we use the publicly available pre-trained weights, we re-train BTS on the complete KITTI depth data set \cite{geiger2012kitti,uhrig2017sparsity} with a depth cap of 100 meters.
%
%
Since the KITTI scene flow data set \cite{geiger2012kitti,menze2015object} provides scene flow labels for the stereo setup in image space, \ie disparity and optical flow displacements, we further transform all estimated depth values for a pixel $\mathbf{p}$ into (virtual) disparity displacements using the available focal length $f$ and baseline $B$ of the stereo camera according to \cref{eq:pinhole}.

\begin{equation} \label{eq:pinhole}
	d_i(\mathbf{p}) = \frac{f \cdot B}{D_i(\mathbf{p})}
\end{equation}

Throughout the remainder of this paper, we denote the (virtual) disparity maps $d_i$ by lowercase letters, and the originally estimated depth map $D_i$ by capital letters.
Anyhow, assuming calibrated cameras, both domains are interchangeable since both provide the necessary geometric 3D information.

\begin{figure}
	\centering
	\begin{subfigure}[c]{0.9\linewidth}
		\includegraphics[width=\linewidth]{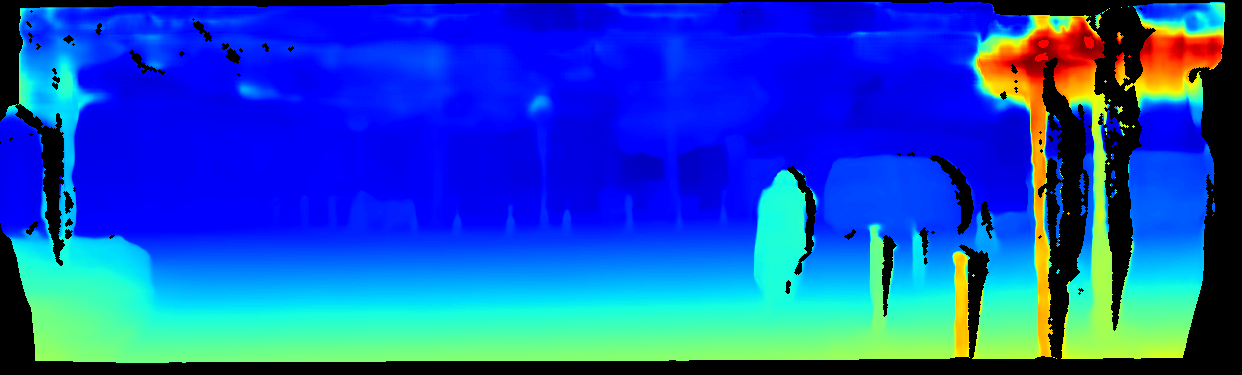}
	\end{subfigure}
	\begin{subfigure}[c]{0.9\linewidth}
		\includegraphics[width=\linewidth]{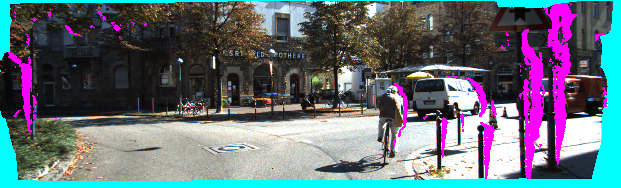}
	\end{subfigure}
	\caption{Illustration of the warping process with occlusion handling. The top images visualizes the geometry (virtual disparity) at time $t_2$ after warping towards the reference frame at time $t_1$. The bottom image shows the two types of obstruction, geometric occlusion (magenta) and out-of-view motion (cyan), which can be considered as occlusion by the camera frustum.}
	\Description{An illustration of gaps in the warped disparity map due to the two types of occlusions.}
	\label{fig:warping}
\end{figure}

\subsection{Warping and Occlusion Estimation} \label{sec:method:warping}
The auxiliary estimators provide the basis for our approach.
However, warping is the actual core and biggest challenge for the combination method.
It is needed to correlate the 3D information of different pixels to find the shift over time.
In accordance with previous work \cite{schuster2018combining}, we define the warping operation to map the current and future depth values as follows:

\begin{equation} \label{eq:warping}
	d^w_2(\mathbf{p}) = d_2\left(\mathbf{p}+\mathbf{u}(\mathbf{p})\right)
\end{equation}

During warping, we ignore target pixels outside of the image domain and use bilinear interpolation to account for sub-pixel displacements in the estimated optical flow.
These out-of-bound regions lead to a non-dense warped disparity.
Additionally, some points move in front or behind others so that not all points visible at $t_1$ are also visible at $t_2$.
These occlusions need to be filtered, because they produce ghosting effects, \ie duplicated objects, after warping.
Since we have an estimate of depth, we can reason about which of these points occlude the others, masking all but the closest point for all source pixels with the same target position.
This is formalized in \cref{eq:occ} by
\begin{equation} \label{eq:occ}
	occ(\mathbf{p}) = \left[ d_1(\mathbf{p}) < d_1(\mathbf{p'}) \forall \mathbf{p'} \in \Omega \mid \mathbf{p'}+\mathbf{u}(\mathbf{p'}) \approx \mathbf{p}+\mathbf{u}(\mathbf{p}) \right]
\end{equation}
to obtain a binary occlusion mask $occ$ for each pixel in the image domain $\Omega$, where $\left[\bullet\right]$ denotes the Iverson bracket and sub-pixel optical flow is handled by rounding.

After initial occlusion mask estimation, we correct discretization and rounding errors by applying two iterations of morphological closing and opening.
An exemplary result of a warped disparity map and occlusion mask is given in \cref{fig:warping}.

By combining the initially estimated optical flow $\mathbf{u}$, the geometric information of $d_1$ at time $t_1$, and the warped disparity $d_2^w$, a non-dense scene flow can be obtained.
This intermediate result is analyzed in \cref{sec:results:ablation}.

\begin{figure}
	\centering
	\begin{subfigure}[c]{0.9\linewidth}
		\includegraphics[width=\linewidth]{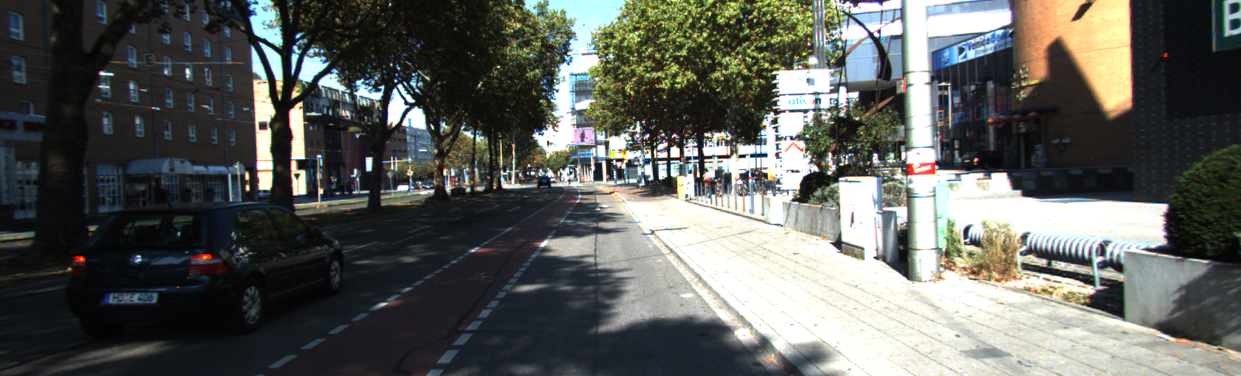}
	\end{subfigure}
	\begin{subfigure}[c]{0.9\linewidth}
		\includegraphics[width=\linewidth]{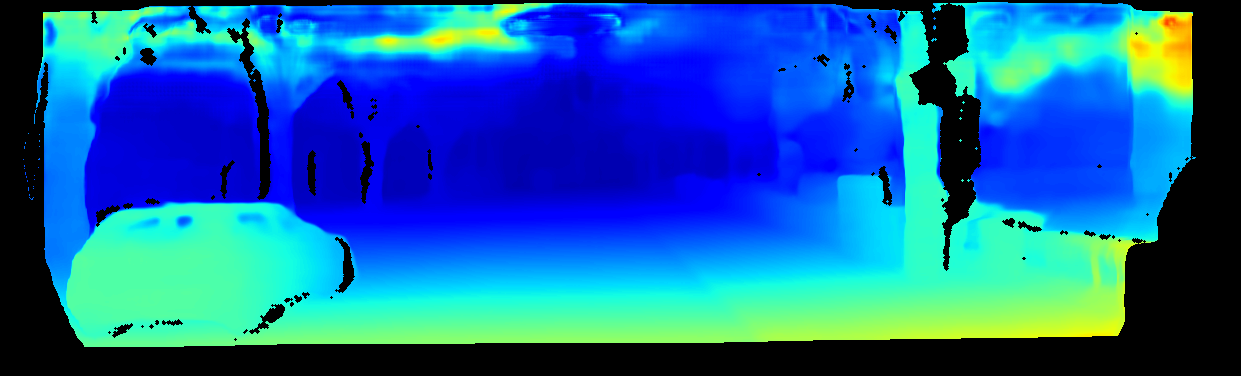}
	\end{subfigure}
	\begin{subfigure}[c]{0.9\linewidth}
		\includegraphics[width=\linewidth]{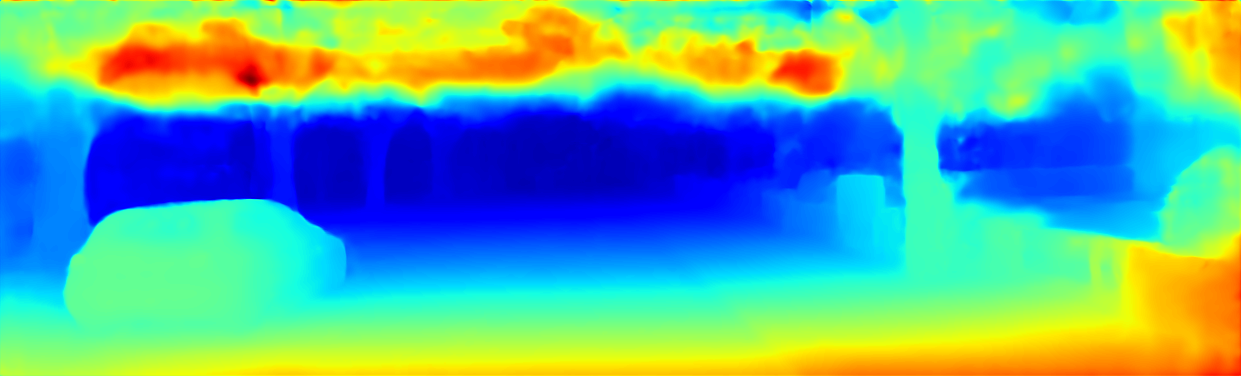}
	\end{subfigure}
	\caption{Visualization of a depth (virtual disparity) estimate from our validation split before and after interpolation with SSGP \cite{schuster2020ssgp}.}
	\Description{The geometry of the scene is visualized before and after interpolation along with a reference image.}
	\label{fig:interpolation}
\end{figure}

\begin{table*}[t]
	\caption{Evaluation of different components and steps of our approach on the validation split of the KITTI scene flow training data. End-point-error (EPE, [px]) and KITTI outlier error (KOE, [\%]) are given. Numbers in parentheses indicate that the respective model was (partially) trained on the validation data.}
	\label{tab:ablation}
	\begin{tabular}{c|cc|cc|cc|cc|c}
	\multirow{2}{*}{Method} & \multicolumn{2}{c|}{D1} & \multicolumn{2}{c|}{D2} & \multicolumn{2}{c|}{OF} & \multicolumn{2}{c|}{SF} & \multirow{2}{*}{Density}\\
	 & EPE & KOE & EPE & KOE & EPE & KOE & $\Sigma$EPE & KOE &\Bstrut\\
	 \hline
	 \hline
	 BTS \cite{lee2019bts} (original) & 3.60 & 24.51 & -- & -- & -- & -- & -- & -- & 100 \%\Tstrut\\
	 BTS (re-trained) & 3.19 & 23.86 & -- & -- & -- & -- & -- & -- & 100 \%\\
	 HD3 \cite{yin2019hierarchical} & -- & -- & -- & -- & (1.74) & (5.13) & -- & -- & 100 \% \\
	 VCN \cite{yang2019volumetric} & -- & -- & -- & -- & (1.41) & (5.00) & -- & -- & 100 \%\Bstrut\\
	 \hline
	 BTS + HD3 &  2.99 & 20.18 & 3.42 & 26.28 & (0.95) & (2.39) & 7.35 & 29.04 & 81.89 \%\Tstrut\\
	 BTS + VCN &  2.98 & 20.15 & 3.36 & 25.44 & (0.76) & (2.70) & 7.10 & 28.07 & 81.70 \%\Bstrut\\
	 \hline
	 BTS + HD3 + SSGP \cite{schuster2020ssgp} (D2) & 3.19 & 23.86 & 3.21 & 24.01 & (1.74) & (5.13) & 8.14 & 35.08 & 100 \%\Tstrut\\
	 BTS + VCN + SSGP (D2) & 3.19 & 23.86 & 3.24 & 24.46 & (1.41) & (5.00) & 7.84 & 35.22 & 100 \%\Bstrut\\
	 \hline
	 BTS + HD3 + SSGP (D1+D2) & 2.76 & 20.30 & 3.21 & 24.01 & (1.74) & (5.13) & 7.70 & 29.50 & 100 \%\Tstrut\\
	 BTS + VCN + SSGP (D1+D2) & 2.76 & 20.30 & 3.24 & 24.46 & (1.41) & (5.00) & 7.41 & 29.34 & 100 \%\Bstrut\\	
	 \hline
	 \hline
	 BTS + SSGP (D1) + OE \cite{yang2020upgrading} & 2.76 & 20.30 & (3.36) & (23.63) & (2.02) & (7.01 ) & 8.14 & 26.53 & 100 \%\Tstrut\\
	 BTS + OE & 3.19 & 23.86 & (3.90) & (27.87) & (2.02) & (7.01 ) & 9.10 & 30.60 & 100 \%\\
	 MonoDepth2 \cite{godard2019digging} + OE & 2.95 & 25.37 & (3.54) & (28.47) & (2.02) & (7.01) & 8.50 & 30.90 & 100 \%\\
	\end{tabular}
\end{table*}

\subsection{Interpolation and Refinement} \label{sec:method:interpolation}
To recover full density, we have to fill in the gaps of the warped disparity $d_2^w$.
Our method of choice is SSGP \cite{schuster2020ssgp}.
SSGP is a recently presented network architecture for image-guided interpolation of different sparse or non-dense information.
It was shown to work for optical flow, scene flow, or depth maps.
We use it to interpolate the gaps in our warped virtual disparity map and name the interpolated output $d_2^i$.
Because of the inverse target domain (disparity instead of depth), we retrain an implementation of SSGP on the KITTI depth data in the disparity space.
This is achieved by conversion of all predicted values and ground truth depth labels with \cref{eq:pinhole} during loss computation.
An example for warped, sparse geometry and the interpolated result is given in \cref{fig:interpolation}.
Note how in this example even the fully occluded bush on the right side is reconstructed reliably by the image guidance of SSGP.

In our experiments in \cref{sec:results:ablation}, we show that the interpolation is able to reconstruct full density (with respect to the image resolution) and additionally to improve the predicted geometric information.
We account this mostly to an correction of the absolute scale of depth (virtual disparity).
This is similar to the two-stage depth estimation in Mono-SF \cite{brickwedde2019mono} where a dedicated network for re-calibration refines the initial depth estimates.

However, based on the observation that SSGP improves the results beyond interpolation, we suggest that this step might also improve already dense input, \ie the (virtual) disparity estimate at time $t_1$.
For this reason, our full model processes the dense estimate $d_1$ with SSGP before combining it into dense scene flow (cf. \cref{fig:overview}). We term the refined disparity map $d_1^r$.

Lastly, all separate results ($\mathbf{u}, d_1^r, d_2^i$) are combined to form dense scene flow (in image space).


\begin{table*}[t]
	\caption{Evaluation results from the KITTI benchmark for all submitted monocular scene flow approaches. We distinguish between supervised and unsupervised methods. Best (lowest) numbers in bold.}
	\label{tab:kitti}
	\begin{tabular}{c|ccc|ccc|ccc|ccc|c}
	\multirow{2}{*}{Method} & \multicolumn{3}{c|}{D1 [\%]} & \multicolumn{3}{c|}{D2 [\%]} & \multicolumn{3}{c|}{OF [\%]} & \multicolumn{3}{c|}{SF [\%]} & \multirow{2}{*}{\begin{tabular}{c}Run\\time\end{tabular}}\\
	 & bg & fg & all & bg & fg & all & bg & fg & all & bg & fg & all &\Bstrut\\
	\hline
	\hline
	 Mono-SF \cite{brickwedde2019mono} & \textbf{14.21}  & 26.94 & \textbf{16.32} & \textbf{16.89} & 33.07 & \textbf{19.59} & 11.40 & 19.64 & 12.77 & \textbf{19.79} & 39.57 & \textbf{23.08} & 41 s\Tstrut\\
	 MonoComb \textbf{(ours)} & 17.89 & \textbf{21.16} & 18.44 & 22.34 & \textbf{25.85} & 22.93 & 5.84 & 8.67 & 6.31 & 27.06 & \textbf{33.55} & 28.14 & 0.58 s \\
	 MonoExpansion \cite{yang2020upgrading}& 24.85 & 27.90 & 25.36 & 27.69 & 31.59 & 28.34 & \textbf{5.83} & \textbf{8.66} & \textbf{6.30} & 29.82 & 36.67 & 30.96 & 0.25 s\\
	Self-Mono-SF-ft \cite{hur2020self} & 20.72 & 29.41 & 22.16 & 23.83 & 32.29 & 25.24 & 15.51 & 17.96 & 15.91 & 31.51 & 45.77 & 33.88 & \textbf{0.09 s}\Bstrut\\
	\hline
	Self-Mono-SF \cite{hur2020self}& 31.22 & 48.04 & 34.02 & 34.89 & 43.59 & 36.34 & 23.26 & 24.93 & 23.54 & 46.68 & 63.82 & 49.54 & \textbf{0.09 s}\Tstrut\\
	\end{tabular}
\end{table*}

\section{Experiments and Results} \label{sec:results}
In our experiments, the monocular combination approach is evaluated step-by-step -- starting from the initial estimates until dense scene flow -- and then compared to state-of-the-art on the KITTI scene flow benchmark \cite{geiger2012kitti,menze2015object}.
This data set provides realistic imagery for diverse traffic scenarios -- including highway, suburban streets, and cities -- as well as labels for full 3D scene flow. 200 sequences are labeled for training and another 200 sequences are provided as input for online evaluation with ground truth labels withhold.
The models for monocular depth estimation are trained with the KITTI depth data set, which is much larger.
These two data sets are related by an intersection of 142 sequences of the labeled training sets.
Therefore, we define our validation split for scene flow as the remaining 58 sequences.
The full list of these sequences is available in the official development kit of the KITTI scene flow data set.

For the evaluation, we consider the default metrics for scene flow as defined by KITTI.
These are the average end-point-errors (EPE) in image space for the separate results, \ie disparities at time $t_1$ (\textit{D1}) and $t_2$ (\textit{D2}) as well as optical flow (\textit{OF}) and the average KITTI outlier rates (KOE), where an estimated value is defined as outlier if its EPE exceeds 3 px or 5 \% of the magnitude of the ground truth value. Scene flow outliers are defined as union of the outliers of separate results, \ie if either disparity values at $t_1$ or $t_2$, or optical flow is an outlier.

\subsection{Ablation Study} \label{sec:results:ablation}
In our ablation study, we evaluate separate parts and intermediate results of our approach individually.
The results are presented in \cref{tab:ablation}.
In detail, we test the re-trained BTS \cite{lee2019bts} with disparity transformation on the scene flow validation set and notice a small improvement over the officially provided pre-trained weights.
Further, the two considered estimators for optical flow, HD3 \cite{yin2019hierarchical} and VCN \cite{yang2019volumetric}, are evaluated.
It is important to note, that we have not re-trained these networks and thus our validation split was entirely used during training of both, HD3 and VCN.
This is indicated by parentheses in \cref{tab:ablation}.
That said, the generalization of the models to the unseen test set is not too bad as is shown  by the results on the KITTI benchmark (cf. \cref{tab:kitti}).

The important next step of our pipeline is the warping.
The warped (virtual) disparities are also evaluated together with the sparse scene flow that is created by warping.
We notice similar performance for both optical flow estimators, reaching densities of 81.9 and 81.7 \%.
It is further evident that the warping introduces some artifacts which reduce the accuracy compared to the disparity maps that are directly estimated in the respective reference frame.
At the same time, the masking of occlusions and out-of-bounds motions also removes some outliers in the disparity at time $t_1$ and the optical flow.
This reveals that HD3 has more difficulties to handle occlusion compared to VCN.
Overall, as seen in previous work \cite{schuster2018combining}, the sparse scene flow obtained by warping is comparatively accurate, yet being non-dense.

To recover full density, $d^w_2$ is interpolated with SSGP.
Interestingly, the interpolation does not only fill in the gaps, but further improves the dense result over the sparse one.
For that reason, we finally apply dense refinement to $d_1$ using SSGP.
This results in a significant reduction of errors for \textit{D1} and an even bigger improvement for the overall scene flow estimates.
At the same time, the qualitative impressions in \cref{fig:interpolation,fig:kitti:ours} reveal that the interpolation and the dense refinement with SSGP introduce artifacts in the sky regions (upper third of the image) where no supervision signal is available in the data set.

Lastly, numbers for the competing method OE \cite{yang2020upgrading} are presented.
This method uses every fifth frame (starting with sequence 0) for validation and the rest for training.
As a consequence, part of our validation set has been used to train OE.
The reference depth estimate for OpticalExpansion was computed with MonoDepth2 \cite{godard2019digging}.
We also report numbers when OE is used in conjunction with BTS and our refined disparity estimate $d_1^r$.
We make this comparison to validate that our approach performs not just better because of the better depth estimator (BTS over MonoDepth2), but because of the way corresponding depth values over time are correlated.
This is validated by the relatively little improvement of scene flow outliers (\textit{SF KOE}) compared to the outliers of \textit{D1} in the last two rows of \cref{tab:ablation}.
The third last row validates that our contribution of dense refinement is also beneficial for other approaches.

%
%
%

\subsection{Comparison to State-of-the-Art} \label{sec:results:sota}
In this experiment we compare our dense monocular scene flow approach to state-of-the-art in this field.
This is done by submitting to the online KITTI scene flow benchmark.
For that, SSGP for interpolation and dense refinement is re-trained on all 200 sequences of the KITTI training split.
This turned out especially useful since our validation split is comparatively large ($> 25 \%$).
Results for all published monocular approaches and our proposed method is shown in \cref{tab:kitti}.
The results of the ablation study (\cref{tab:ablation}) are transferred within reasonable deviation and some further improvements due to the additional training data.

Our monocular combination approach (MonoComb) pushes state-of-art in various ways.
We achieve the second best result overall and the best result among all methods with sub-second run time.
Further, we achieve the lowest error rate for the important foreground regions (\textit{fg}) of dynamic objects with a margin of more than 3 percentage points.

A qualitative comparison is provided in \cref{fig:kitti} where the result of the first test frame for all monocular methods is visualized along with the corresponding error maps.
This particular sample reveals the major issues of each approach.
For Mono-SF \cite{brickwedde2019mono} the biggest challenge is the correct estimation of dynamic objects.
This is also reflected by the quantitative results in \cref{tab:kitti}.
The monocular version of OpticalExpansion \cite{yang2020upgrading} (MonoExpansion), is depending a lot on the estimate of the initial depth/disparity. Further, since the relative change in depth is tightly coupled and estimated together with optical flow, the estimated second disparity suffers from the same limitations as optical flow, \ie occlusions, visual perturbation by large geometric deformations over time, etc.
The (fine-tuned) self-supervised approach \cite{hur2020self} is mainly restricted by the lower level of details due to the reduced output resolution.
Our proposed approach predicts scene flow at full resolution, handles occlusion explicitly and solves this issue by interpolation, and provides top performance for dynamic objects.
However, the error maps in \cref{fig:kitti:ours} indicate that the absolute scale is not recovered sufficiently well.
Further, the major limitation of our method is the inconsistency of the prediction due to the  separation. 
This results in bad alignment of correct and erroneous regions across the separate tasks, and ultimately in a high outlier rate in the scene flow metric.
In fact in \cref{tab:kitti}, our approach has a much larger margin between the highest outlier rate of \textit{D1}, \textit{D2}, or \textit{OF} and \textit{SF}, compared to \eg OpticalExpansion \cite{yang2020upgrading}.

\begin{figure*}[!b]
	\newcommand{\VizFigureWidth}{0.4\linewidth}
	\centering
	\begin{subfigure}[c]{\VizFigureWidth}
		\centering
		\includegraphics[width=\linewidth]{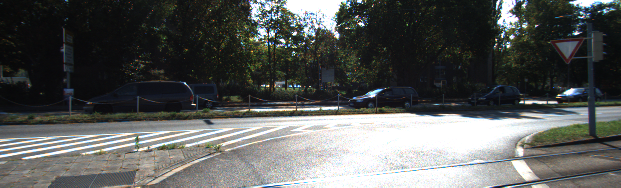}\\%
		\caption{Reference Image}
		\label{fig:kitti:img}
		\vspace{3mm}
	\end{subfigure}
	
	\begin{subfigure}[c]{\linewidth}
		\begin{center}
			\begin{subfigure}[c]{\VizFigureWidth}
				\centering
				Estimate
			\end{subfigure}
			\begin{subfigure}[c]{\VizFigureWidth}
				\centering
				Error Map
			\end{subfigure}\\%
			\begin{subfigure}[c]{0.03\linewidth}
				\centering
				\rotatebox[origin=c]{90}{D1}
			\end{subfigure}
			\begin{subfigure}[c]{\VizFigureWidth}
				\includegraphics[width=\linewidth]{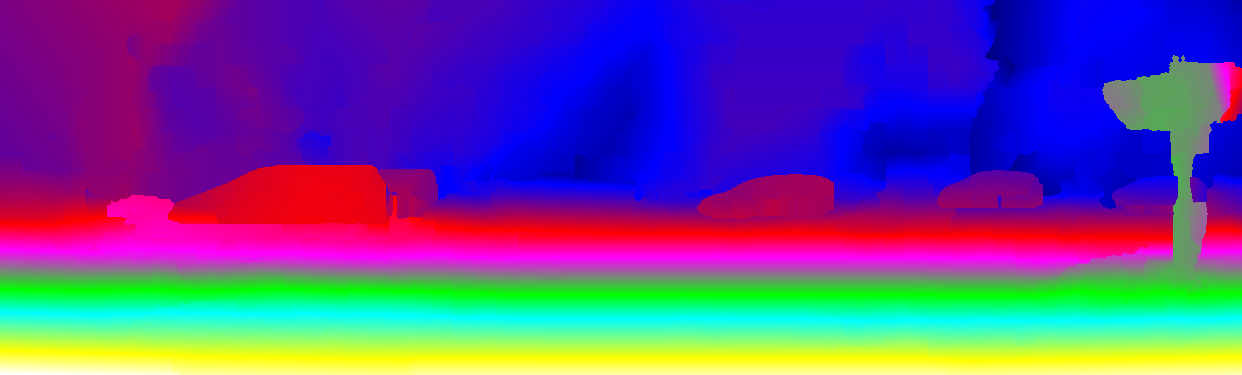}
			\end{subfigure}
			\begin{subfigure}[c]{\VizFigureWidth}
				\includegraphics[width=\linewidth]{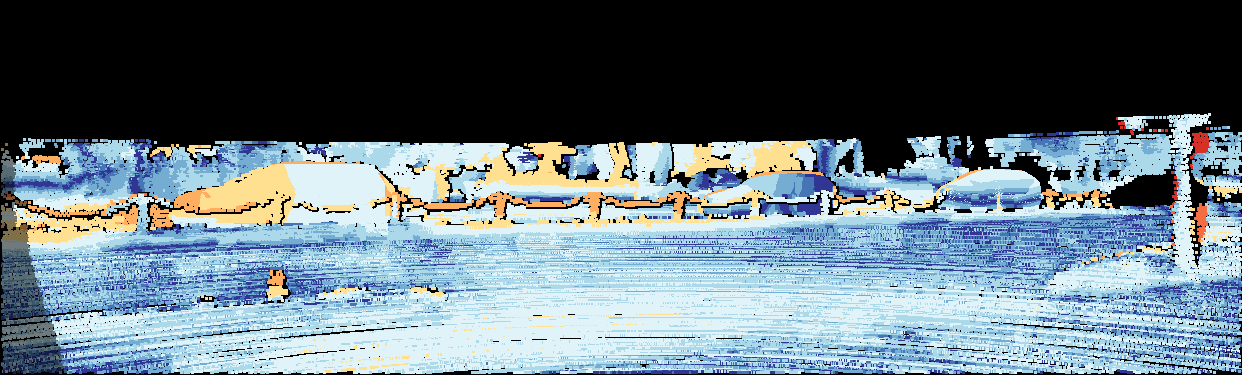}
			\end{subfigure}
			\begin{subfigure}[c]{0.03\linewidth}
				\centering
				\rotatebox[origin=c]{90}{D1}
			\end{subfigure}\\%
			\begin{subfigure}[c]{0.03\linewidth}
				\centering
				\rotatebox[origin=c]{90}{D2}
			\end{subfigure}
			\begin{subfigure}[c]{\VizFigureWidth}
				\includegraphics[width=\linewidth]{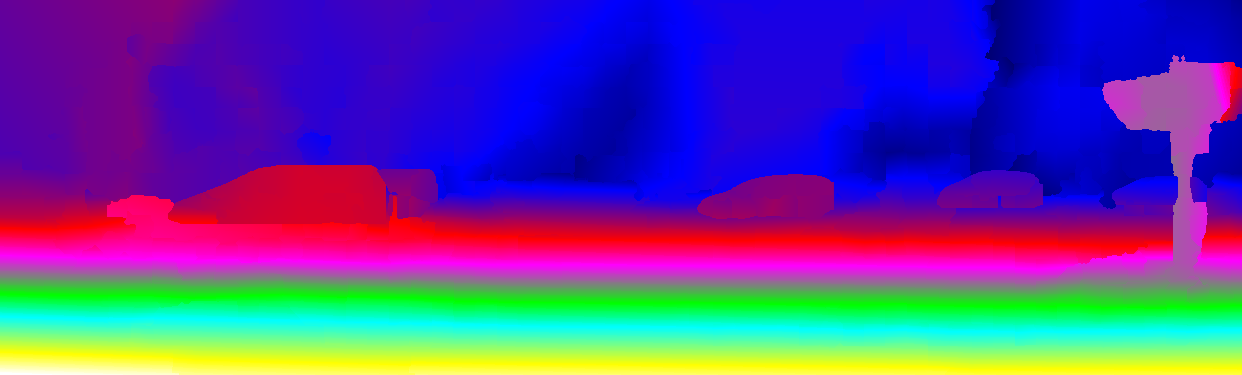}
			\end{subfigure}
			\begin{subfigure}[c]{\VizFigureWidth}
				\includegraphics[width=\linewidth]{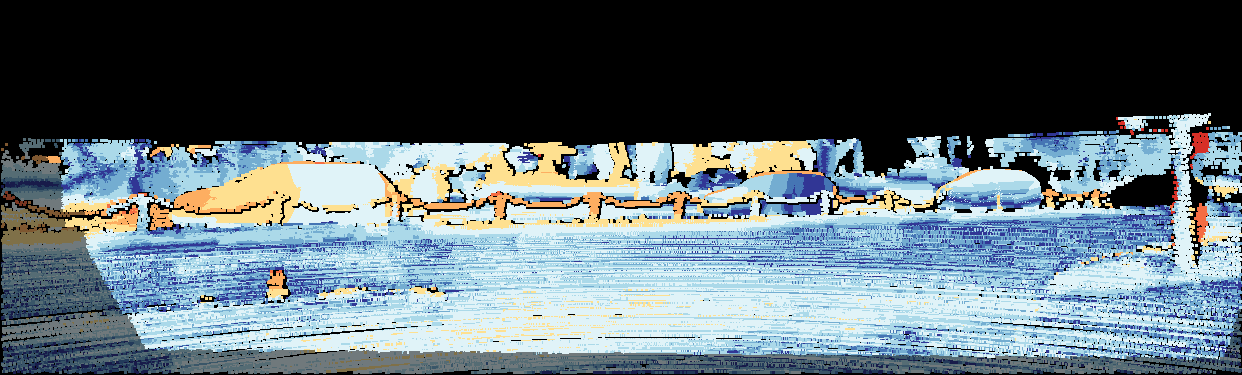}
			\end{subfigure}
			\begin{subfigure}[c]{0.03\linewidth}
				\centering
				\rotatebox[origin=c]{90}{D2}
			\end{subfigure}\\%
			\begin{subfigure}[c]{0.03\linewidth}
				\centering
				\rotatebox[origin=c]{90}{OF}
			\end{subfigure}
			\begin{subfigure}[c]{\VizFigureWidth}
				\includegraphics[width=\linewidth]{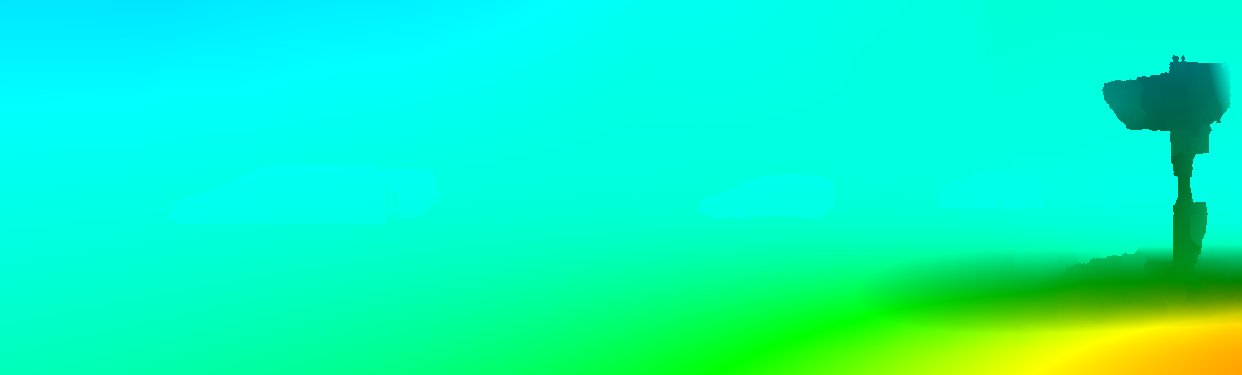}
			\end{subfigure}
			\begin{subfigure}[c]{\VizFigureWidth}
				\includegraphics[width=\linewidth]{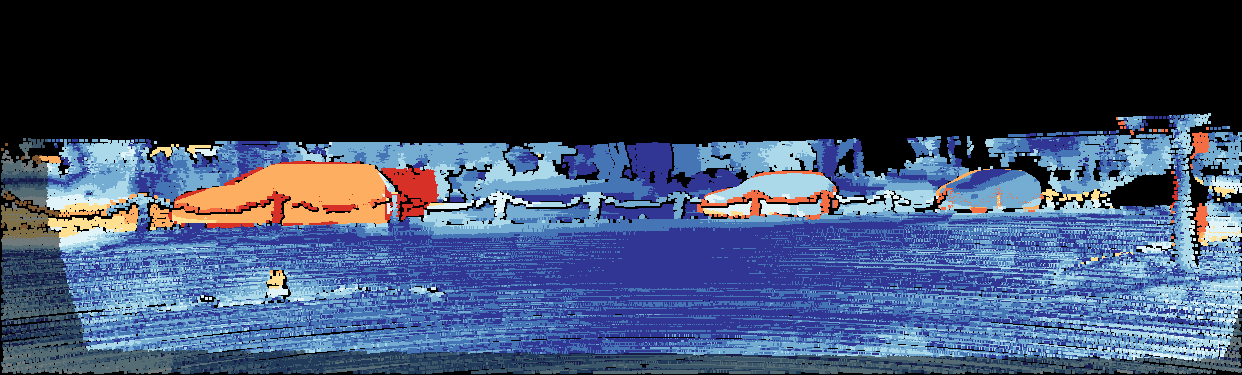}
			\end{subfigure}
			\begin{subfigure}[c]{0.03\linewidth}
				\centering
				\rotatebox[origin=c]{90}{OF}
			\end{subfigure}\\%
			\begin{subfigure}[c]{0.04\linewidth}
				\centering
				EPE:
			\end{subfigure}
			\begin{subfigure}[c]{0.8\linewidth}
				\includegraphics[width=\linewidth]{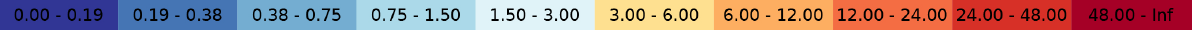}
			\end{subfigure}
			\begin{subfigure}[c]{0.04\linewidth}
				\Bstrut
			\end{subfigure}\\%
		\end{center}
		\caption{Result of Mono-SF \cite{brickwedde2019mono}}
		\label{fig:kitti:monosf}
		\vspace{3mm}
	\end{subfigure}
	
	\begin{subfigure}[c]{\linewidth}
		\begin{center}
			\begin{subfigure}[c]{\VizFigureWidth}
				\centering
				Estimate
			\end{subfigure}
			\begin{subfigure}[c]{\VizFigureWidth}
				\centering
				Error Map
			\end{subfigure}\\%
			\begin{subfigure}[c]{0.03\linewidth}
				\centering
				\rotatebox[origin=c]{90}{D1}
			\end{subfigure}
			\begin{subfigure}[c]{\VizFigureWidth}
				\includegraphics[width=\linewidth]{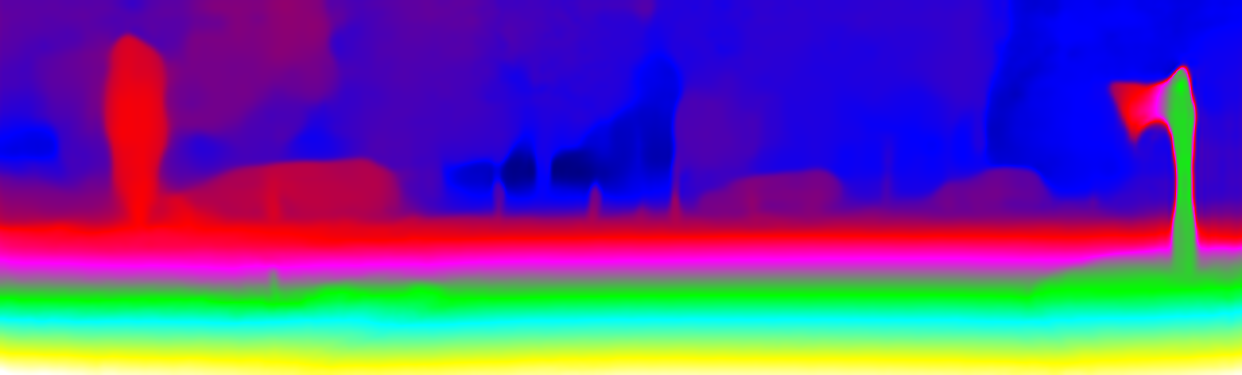}
			\end{subfigure}
			\begin{subfigure}[c]{\VizFigureWidth}
				\includegraphics[width=\linewidth]{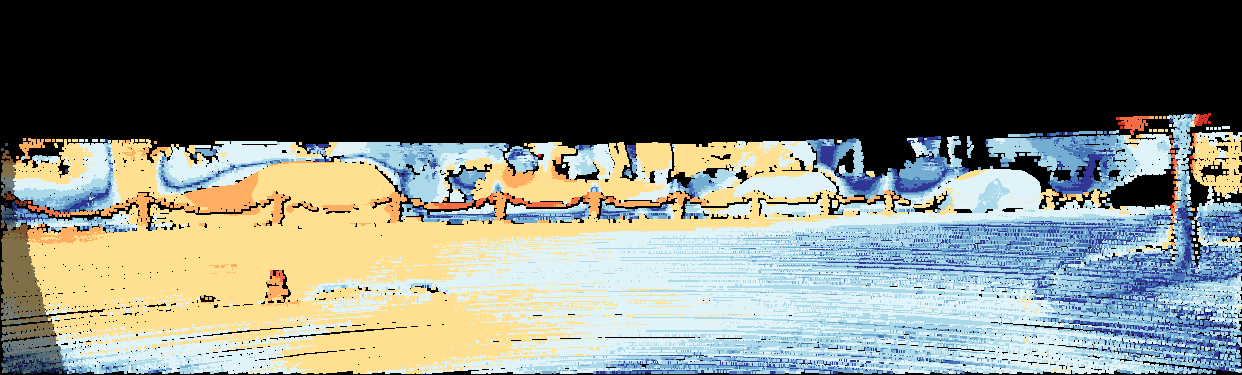}
			\end{subfigure}
			\begin{subfigure}[c]{0.03\linewidth}
				\centering
				\rotatebox[origin=c]{90}{D1}
			\end{subfigure}\\%
			\begin{subfigure}[c]{0.03\linewidth}
				\centering
				\rotatebox[origin=c]{90}{D2}
			\end{subfigure}
			\begin{subfigure}[c]{\VizFigureWidth}
				\includegraphics[width=\linewidth]{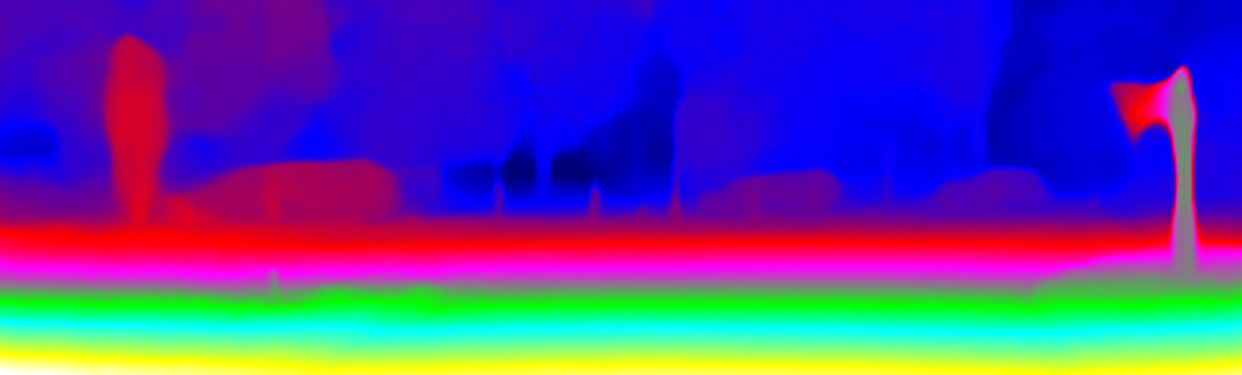}
			\end{subfigure}
			\begin{subfigure}[c]{\VizFigureWidth}
				\includegraphics[width=\linewidth]{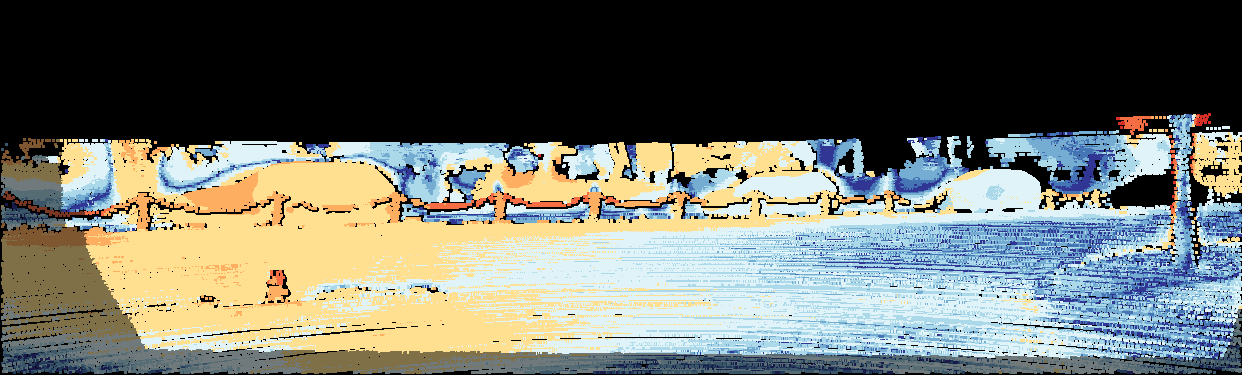}
			\end{subfigure}
			\begin{subfigure}[c]{0.03\linewidth}
				\centering
				\rotatebox[origin=c]{90}{D2}
			\end{subfigure}\\%
			\begin{subfigure}[c]{0.03\linewidth}
				\centering
				\rotatebox[origin=c]{90}{OF}
			\end{subfigure}
			\begin{subfigure}[c]{\VizFigureWidth}
				\includegraphics[width=\linewidth]{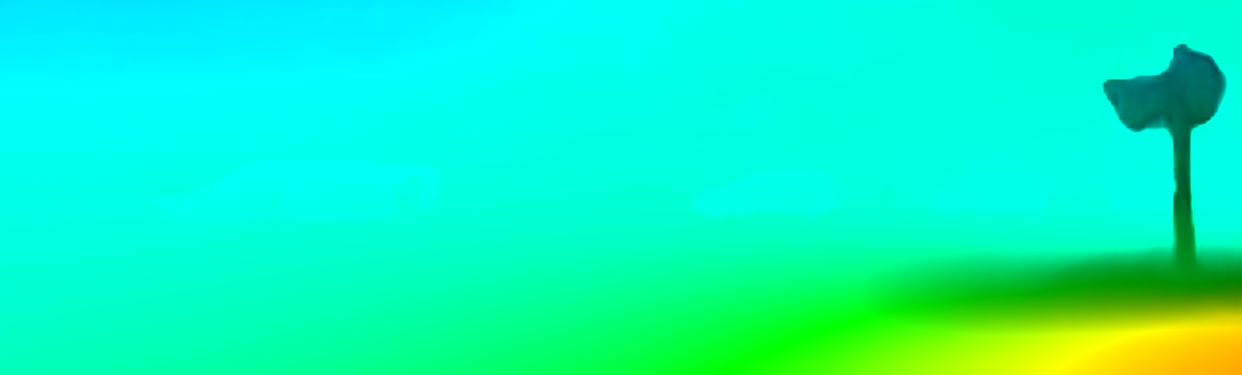}
			\end{subfigure}
			\begin{subfigure}[c]{\VizFigureWidth}
				\includegraphics[width=\linewidth]{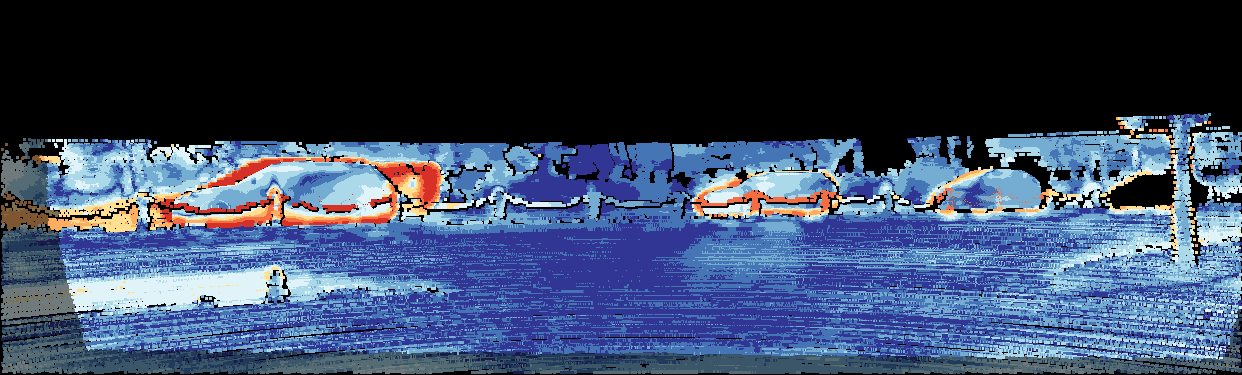}
			\end{subfigure}
			\begin{subfigure}[c]{0.03\linewidth}
				\centering
				\rotatebox[origin=c]{90}{OF}
			\end{subfigure}\\%
			\begin{subfigure}[c]{0.04\linewidth}
				\centering
				EPE:
			\end{subfigure}
			\begin{subfigure}[c]{0.8\linewidth}
				\includegraphics[width=\linewidth]{img/errorbar}
			\end{subfigure}
			\begin{subfigure}[c]{0.04\linewidth}
				\Bstrut
			\end{subfigure}\\%
		\end{center}
		\caption{Result of MonoExpansion \cite{yang2020upgrading}}
		\label{fig:kitti:expansion}
	\end{subfigure}
	
	\Description{For all compared monocular scene flow approaches, a visualization of the results along with the corresponding error maps.}
	\caption{Continued on the next page.}
\end{figure*}

\begin{figure*}[t]
	\ContinuedFloat
	\newcommand{\VizFigureWidth}{0.4\linewidth}
	\begin{subfigure}[c]{\linewidth}
		\begin{center}
			\begin{subfigure}[c]{\VizFigureWidth}
				\centering
				Estimate
			\end{subfigure}
			\begin{subfigure}[c]{\VizFigureWidth}
				\centering
				Error Map
			\end{subfigure}\\%
			\begin{subfigure}[c]{0.03\linewidth}
				\centering
				\rotatebox[origin=c]{90}{D1}
			\end{subfigure}
			\begin{subfigure}[c]{\VizFigureWidth}
				\includegraphics[width=\linewidth]{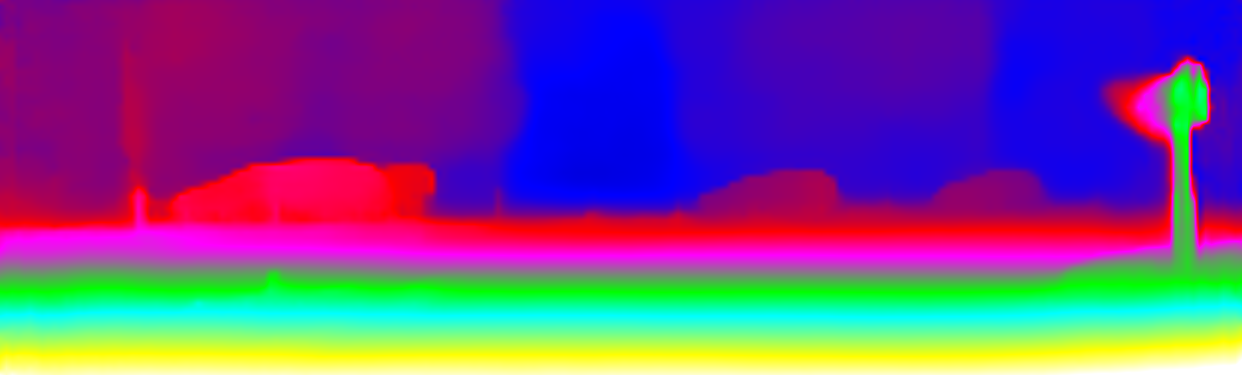}
			\end{subfigure}
			\begin{subfigure}[c]{\VizFigureWidth}
				\includegraphics[width=\linewidth]{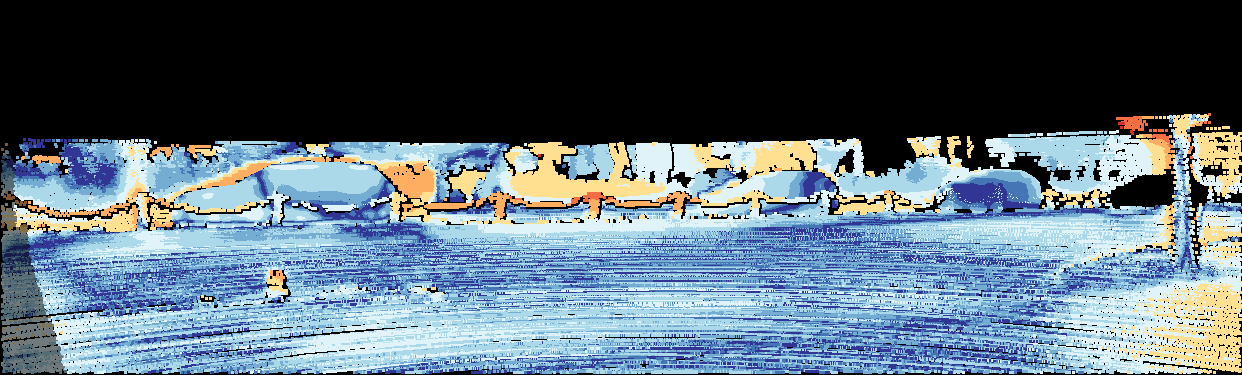}
			\end{subfigure}
			\begin{subfigure}[c]{0.03\linewidth}
				\centering
				\rotatebox[origin=c]{90}{D1}
			\end{subfigure}\\%
			\begin{subfigure}[c]{0.03\linewidth}
				\centering
				\rotatebox[origin=c]{90}{D2}
			\end{subfigure}
			\begin{subfigure}[c]{\VizFigureWidth}
				\includegraphics[width=\linewidth]{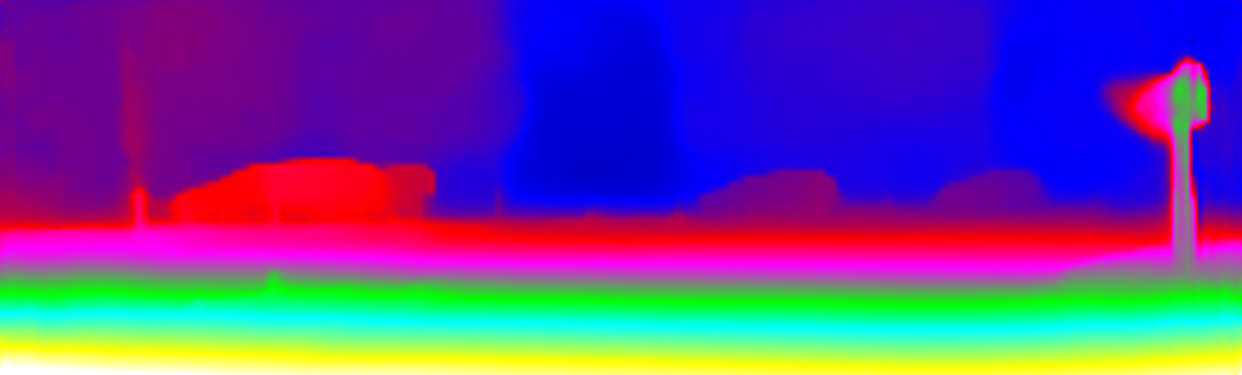}
			\end{subfigure}
			\begin{subfigure}[c]{\VizFigureWidth}
				\includegraphics[width=\linewidth]{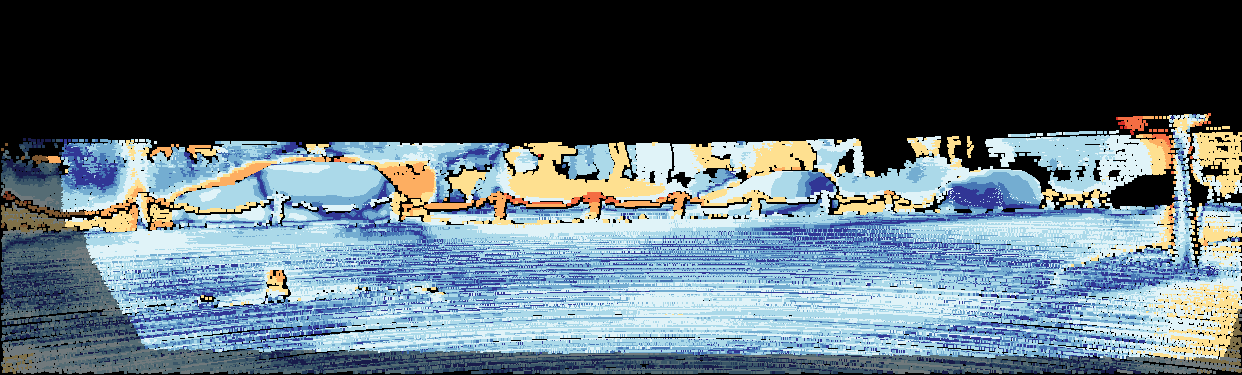}
			\end{subfigure}
			\begin{subfigure}[c]{0.03\linewidth}
				\centering
				\rotatebox[origin=c]{90}{D2}
			\end{subfigure}\\%
			\begin{subfigure}[c]{0.03\linewidth}
				\centering
				\rotatebox[origin=c]{90}{OF}
			\end{subfigure}
			\begin{subfigure}[c]{\VizFigureWidth}
				\includegraphics[width=\linewidth]{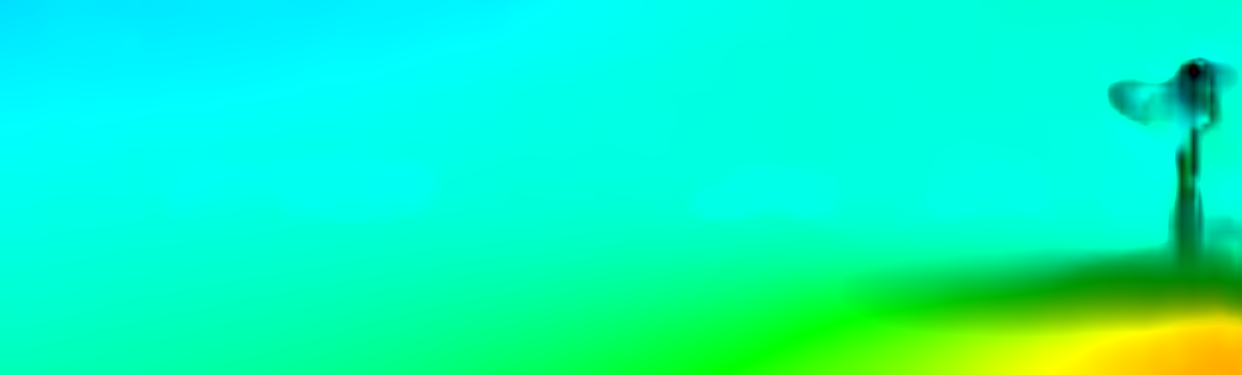}
			\end{subfigure}
			\begin{subfigure}[c]{\VizFigureWidth}
				\includegraphics[width=\linewidth]{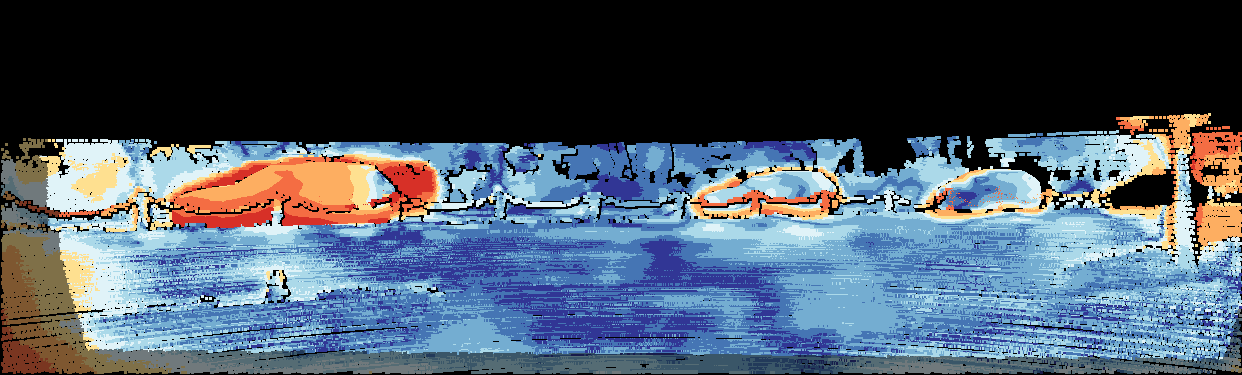}
			\end{subfigure}
			\begin{subfigure}[c]{0.03\linewidth}
				\centering
				\rotatebox[origin=c]{90}{OF}
			\end{subfigure}\\%
			\begin{subfigure}[c]{0.04\linewidth}
				\centering
				EPE:
			\end{subfigure}
			\begin{subfigure}[c]{0.8\linewidth}
				\includegraphics[width=\linewidth]{img/errorbar}
			\end{subfigure}
			\begin{subfigure}[c]{0.04\linewidth}
				\Bstrut
			\end{subfigure}\\%
		\end{center}
		\caption{Result of Self-Mono-SF (fine-tuned) \cite{hur2020self}}
		\label{fig:kitti:selfmonosf}
		\vspace{3mm}
	\end{subfigure}
	
	\begin{subfigure}[c]{\linewidth}
		\begin{center}
			\begin{subfigure}[c]{\VizFigureWidth}
				\centering
				Estimate
			\end{subfigure}
			\begin{subfigure}[c]{\VizFigureWidth}
				\centering
				Error Map
			\end{subfigure}\\%
			\begin{subfigure}[c]{0.03\linewidth}
				\centering
				\rotatebox[origin=c]{90}{D1}
			\end{subfigure}
			\begin{subfigure}[c]{\VizFigureWidth}
				\includegraphics[width=\linewidth]{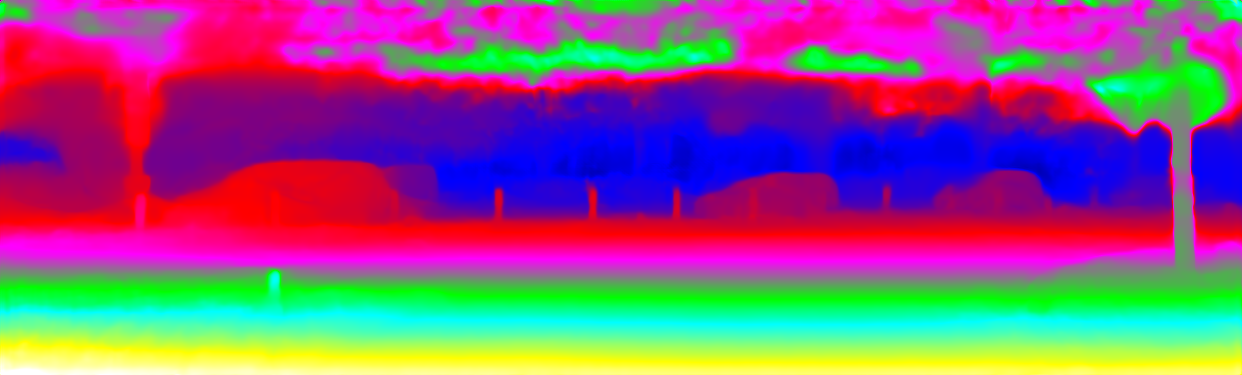}
			\end{subfigure}
			\begin{subfigure}[c]{\VizFigureWidth}
				\includegraphics[width=\linewidth]{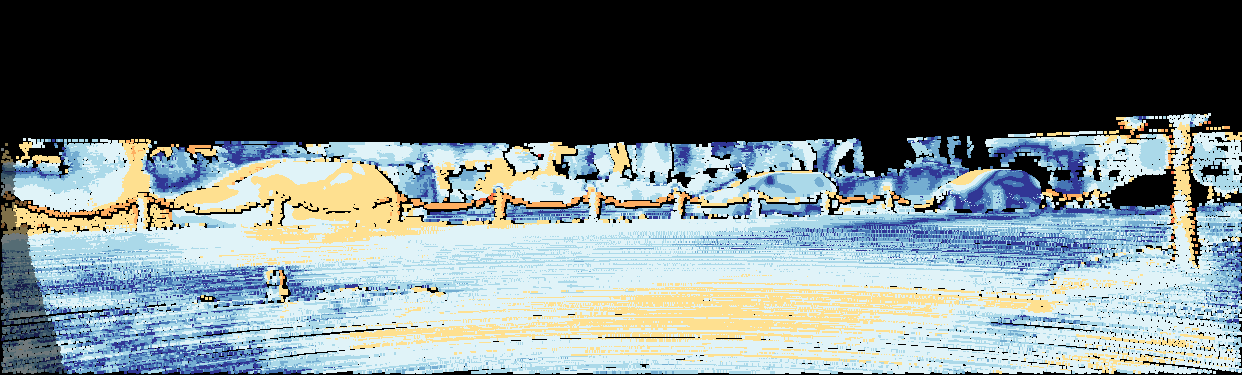}
			\end{subfigure}
			\begin{subfigure}[c]{0.03\linewidth}
				\centering
				\rotatebox[origin=c]{90}{D1}
			\end{subfigure}\\%
			\begin{subfigure}[c]{0.03\linewidth}
				\centering
				\rotatebox[origin=c]{90}{D2}
			\end{subfigure}
			\begin{subfigure}[c]{\VizFigureWidth}
				\includegraphics[width=\linewidth]{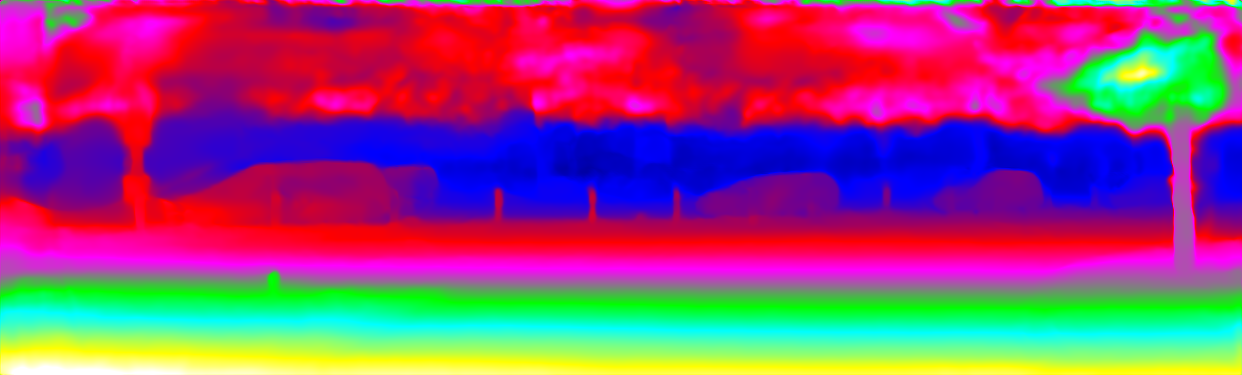}
			\end{subfigure}
			\begin{subfigure}[c]{\VizFigureWidth}
				\includegraphics[width=\linewidth]{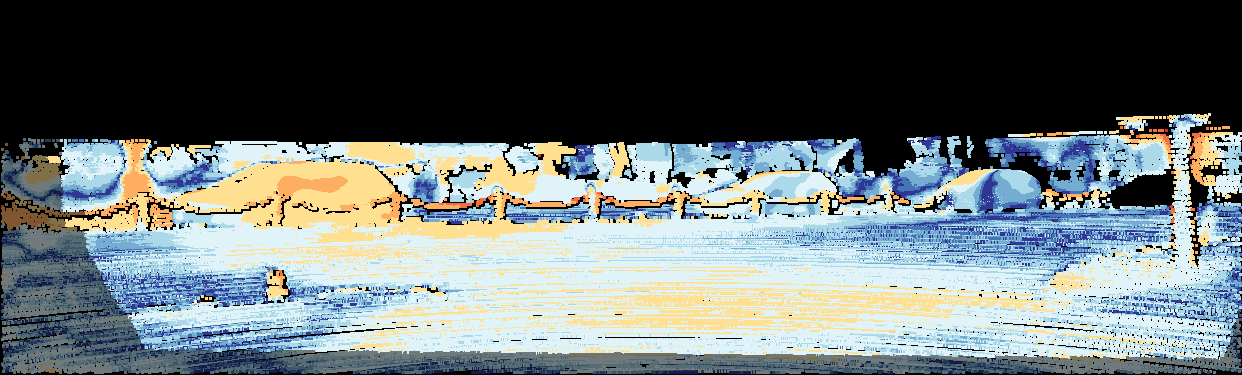}
			\end{subfigure}
			\begin{subfigure}[c]{0.03\linewidth}
				\centering
				\rotatebox[origin=c]{90}{D2}
			\end{subfigure}\\%
			\begin{subfigure}[c]{0.03\linewidth}
				\centering
				\rotatebox[origin=c]{90}{OF}
			\end{subfigure}
			\begin{subfigure}[c]{\VizFigureWidth}
				\includegraphics[width=\linewidth]{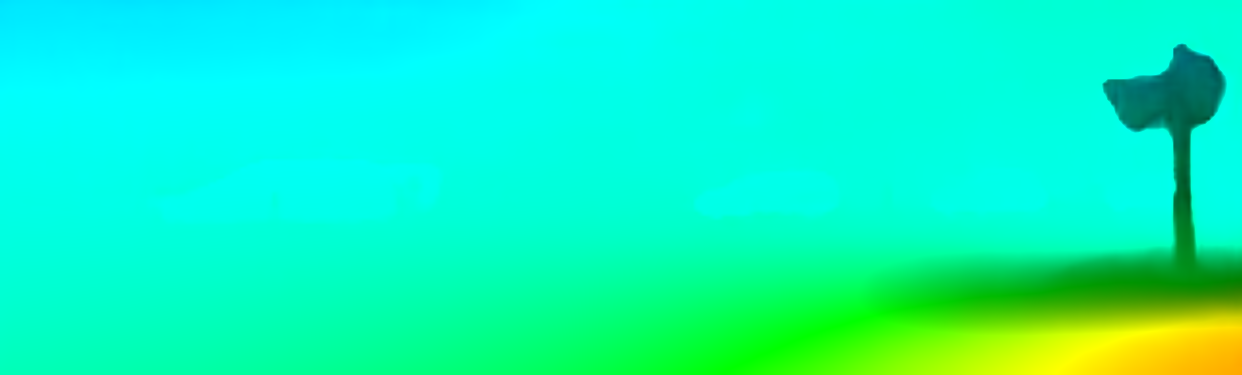}
			\end{subfigure}
			\begin{subfigure}[c]{\VizFigureWidth}
				\includegraphics[width=\linewidth]{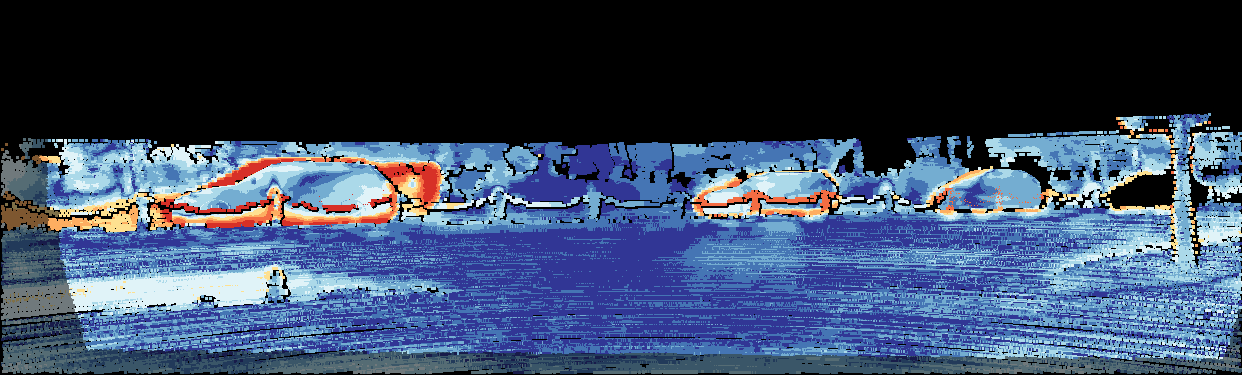}
			\end{subfigure}
			\begin{subfigure}[c]{0.03\linewidth}
				\centering
				\rotatebox[origin=c]{90}{OF}
			\end{subfigure}\\%
			\begin{subfigure}[c]{0.04\linewidth}
				\centering
				EPE:
			\end{subfigure}
			\begin{subfigure}[c]{0.8\linewidth}
				\includegraphics[width=\linewidth]{img/errorbar}
			\end{subfigure}
			\begin{subfigure}[c]{0.04\linewidth}
				\Bstrut
			\end{subfigure}\\%
		\end{center}
		\caption{Result of our monocular combination approach MonoComb.}
		\label{fig:kitti:ours}
	\end{subfigure}
	
	\caption{Visualization and error maps from the KITTI online benchmark for the first test frame. We compare Mono-SF \cite{brickwedde2019mono} (\subref{fig:kitti:monosf}), MonoExpansion \cite{yang2020upgrading} (\subref{fig:kitti:expansion}), Self-Mono-SF \cite{hur2020self} (\subref{fig:kitti:selfmonosf}), and our proposed approach (\subref{fig:kitti:ours}).} 
	\Description{For all compared monocular scene flow approaches, a visualization of the results along with the corresponding error maps.}
	\label{fig:kitti}
\end{figure*}

\subsection{Run Time} \label{sec:results:runtime}
One advantage of the combination approach is the modularity.
As part of that, the overall run time is defined by the sum of the auxiliary run times as given by \cref{tab:runtime}.
In our case this leads to an approximate average run time per frame of 0.58 seconds.
This sums up over two runs of the single image depth estimator, one forward pass of the optical flow estimator, and two runs of SSGP for refinement and interpolation.
The time for warping and combination of the separate results is neglectable small.

\begin{table}[!h]
	\caption{Breakdown of the run time.}
	\label{tab:runtime}
	\begin{tabular}{c|c|c}
		Module & Calls & Run time\Bstrut\\
		\hline
		HD3 \cite{yin2019hierarchical} / VCN \cite{yang2019volumetric} & 1 & 0.1 / 0.18 s\Tstrut\\
		BTS \cite{lee2019bts} & 2 & 0.06 s\\
		SSGP \cite{schuster2020ssgp} & 2 & 0.14 s\Bstrut\\
		\hline
		Total & -- & 0.5 / 0.58 s\Tstrut\\ 
	\end{tabular}
\end{table}

\section{Conclusion} \label{sec:conclusion}
The sparse-to-dense re-combination approach for scene flow estimation is successfully transferred to the monocular camera setup.
This is achieved by latest success in single image depth estimation and robust sparse-to-dense interpolation.
Together with state-of-the-art auxiliary estimators, the proposed concept achieves more than competitive results at reasonable speed.
The modularity of the approach allows to quickly replace parts of the pipeline to tune the overall performance towards a certain goal, \ie real time performance.

In this relatively new discipline of monocular scene flow estimation, the monocular combination approach can be seen as a strong baseline for future developments.

The biggest limitation of our method is the inconsistency of the separate results.
To overcome this in the future, we propose to perform interpolation and dense refinement jointly, which might also introduce mutual advantages for both tasks.
Additionally, the monocular camera setup for scene flow estimation does not restrict the depth estimation to use a single image.
It should be investigated whether the joint estimation of depth over time (\eg two-view depth estimation) can improve the results further.

\begin{acks}
This work was partially funded by the BMW Group and partially by the Federal Ministry of Education and Research Germany under the project VIDETE (01IW18002).
\end{acks}

\balance
\bibliographystyle{ACM-Reference-Format}
\bibliography{bib}

\end{document}